\documentclass{ieeeaccess}
\usepackage{cite}
\usepackage{amsmath,amssymb,amsfonts}
\usepackage{algorithmic}
\usepackage{graphicx}
\usepackage{textcomp}

\usepackage{bm}
\makeatletter
\AtBeginDocument{\DeclareMathVersion{bold}
\SetSymbolFont{operators}{bold}{T1}{times}{b}{n}
\SetSymbolFont{NewLetters}{bold}{T1}{times}{b}{it}
\SetMathAlphabet{\mathrm}{bold}{T1}{times}{b}{n}
\SetMathAlphabet{\mathit}{bold}{T1}{times}{b}{it}
\SetMathAlphabet{\mathbf}{bold}{T1}{times}{b}{n}
\SetMathAlphabet{\mathtt}{bold}{OT1}{pcr}{b}{n}
\SetSymbolFont{symbols}{bold}{OMS}{cmsy}{b}{n}
\renewcommand\boldmath{\@nomath\boldmath\mathversion{bold}}}
\makeatother

\def\BibTeX{{\rm B\kern-.05em{\sc i\kern-.025em b}\kern-.08em
    T\kern-.1667em\lower.7ex\hbox{E}\kern-.125emX}}

\begin{document}
\history{Date of publication xxxx 00, 0000, date of current version xxxx 00, 0000.}
\doi{10.1109/ACCESS.2023.1120000}

\title{The Promise of Analog Deep Learning: Recent Advances, Challenges and Opportunities}
\author{\uppercase{Aditya Datar*}\authorrefmark{1}, \uppercase{Pramit Saha*}\authorrefmark{2}}
\address[1]{Independent Researcher}\address[2]{University of Oxford}
\tfootnote{* Indicates Equal Contribution}

\markboth
{Datar \headeretal: The Promise of Analog Deep Learning}
{Datar \headeretal: The Promise of Analog Deep Learning}

\corresp{Corresponding author: Pramit Saha (e-mail: pramit.saha@eng.ox.ac.uk).}

\begin{abstract}
Much of the present-day Artificial Intelligence (AI) utilizes artificial neural networks, which are sophisticated computational models designed to recognize patterns and solve complex problems by learning from data. However, a major bottleneck occurs during a device’s calculation of weighted sums for forward propagation and optimization procedure for backpropagation, especially for deep neural networks, or networks with numerous layers. Exploration into different methods of implementing neural networks is necessary for further advancement of the area. While a great deal of research into AI hardware in both directions, analog and digital implementation widely exists, much of the existing survey works lacks discussion on the progress of analog deep learning. To this end, we attempt to evaluate and specify the advantages and disadvantages, along with the current progress with regards to deep learning, for analog implementations. In this paper, our focus lies on the comprehensive examination of eight distinct analog deep learning methodologies across multiple key parameters. These parameters include attained accuracy levels, application domains, algorithmic advancements, computational speed, and considerations of energy efficiency and power consumption. We also identify the neural network-based experiments implemented using these hardware devices and discuss comparative performance achieved by the different analog deep learning methods along with an analysis of their current limitations. Overall, we find that Analog Deep Learning has great potential for future consumer-level applications, but there is still a long road ahead in terms of scalability. Most of the current implementations are more proof of concept and are not yet practically deployable for large-scale models.
\end{abstract}


\titlepgskip=-21pt

\maketitle
\section{Introduction}
\IEEEPARstart{W}{ith} the advent of Artificial Intelligence, Deep Learning has enabled significantly greater capabilities by adding numerous layers to Artificial Neural Networks (ANNs) \cite{cation3, cation4, cation6, ferroelectric6, pc6, sc2, spin2}. These additional layers aim to mimic the human brain more closely, which comprises an estimated 10 billion neurons, each possessing capabilities far beyond those of artificial neurons present in ANNs. While artificial neurons operate on abstract activation functions, human neurons are influenced by a diverse array of factors, including time, space, the strength of synaptic connections, and electrical pulses. Although it remains challenging to simulate all aspects of human decision-making, increasing the number of neurons and synaptic connections in ANNs through additional layers brings us closer to mimicking the human brain's behavior. Consequently, the possibilities for AI applications become immensely vast, and employing larger neural networks emerges as a promising approach to realize these potentials.
 
In an ANN, each neuron establishes connections with every other neuron through weighted synapses. These connections allow a neuron to transmit a value scaled by the weight of the synapse, contributing to the activation of other neurons. These "multiply-and-add" operations can be simplified as matrix multiplications, where the neuron values form a matrix that gets multiplied by the matrix of synapse weights. This approach facilitates more straightforward code implementations using matrix libraries. Present-day Artificial Intelligence heavily relies on digital-oriented hardware, particularly Graphics Processing Units (GPUs), to perform binary matrix multiplications in neural networks. However, it is worth noting that a single binary matrix multiplication involves a significantly large number of steps due to the nature of binary computations. As a result, digital hardware does not offer the most efficient pathway to building much larger networks without compromising on time, space, and energy efficiency. Therefore, to expand Deep Learning to encompass hundreds of layers while maintaining practicality, we must explore alternatives to digital implementations of ANNs that can provide both accuracy and efficiency.

The objective of this review is to provide a comprehensive overview and comparison of different emerging technologies for Analog Deep Learning. These technologies aim to enhance AI capabilities by offering efficient and accurate alternatives to the existing digital implementations of ANNs. By exploring the potential of Analog Deep Learning, we seek to enable the practical realization of significantly larger neural networks, paving the way for further advancements in the field of Artificial Intelligence. Although previous works addressing the Analog Deep Learning are focused on the benefits of Analog techniques for applications in deep learning, we hope to establish a baseline review of the area, evaluating both the benefits and consequences to determine whether Analog Deep Learning has great potential and should continue to be studied for future implementations. To the best of our knowledge, this is the first work that presents a detailed overview of analog deep learning and its progress from a computational point of view. 

In this context, our primary contributions are two-fold:
\begin{enumerate}
    \item Provide an in-depth comparison of existing Analog Deep Learning techniques in terms of algorithms implemented, performance achieved, speed, and power consumption.
    \item Analyze the current state and future prospects for Analog Deep Learning
\end{enumerate}

The rest of the paper is organised as follows. In Section II, we introduce preliminary concepts like Deep Learning, Analog Deep learning, and Neuromorphic Computing. In Section III, we define and explain the different analog deep learning implementations along with a visual representation of the taxonomy. Section IV elaborates and compares the aforementioned analog strategies in terms of three main evaluation parameters - (i) Algorithms, Applications, and Accuracy, (ii) Computational Speed, and (iii) Energy Efficiency and Power Consumption. Next, we discuss the insights derived from the comparison and recommendations based on the insights in Section V. Finally, we conclude the article in Section VI.
\section{Background}

\subsection{Deep Learning}
Deep learning is a subfield of machine learning that involves training artificial neural networks to learn and make predictions from data. These neural networks, inspired by the structure of the human brain, consist of layers of interconnected nodes that process and transform input data to generate meaningful outputs. Deep learning excels in tasks like image and speech recognition, natural language processing, and decision-making by automatically learning relevant features and patterns from large datasets, enabling the network to generalize and perform well on new, unseen datasets\cite{cation1, cation2, cation4, anion3, anion5, ferroelectric6, pc6, spin2, spin3, spin4}.

Artificial Neural Networks have enabled us to imitate certain biological brain functions; however, achieving capabilities closely resembling the human brain demands the adoption of deep neural networks, typically comprising 2 or more hidden layers. To even approach human-level decision-making, ANNs with hundreds of hidden layers are imperative. In the quest to make the fully connected ANNs more efficient, effective, and scalable, various advanced neural network alternatives have been proposed, including Convolutional Neural Networks (CNNs) \cite{anion5, pc1, spin5, yao2020fully}, Spiking Neural Networks (SNNs) \cite{pc5, mit2}, Recurrent Neural Networks (RNNs) \cite{cation1}, and Transformer Neural Networks (TNNs).
\par
Artificial Neural Networks serve as simple feed-forward neural networks applicable to a wide range of Artificial Intelligence problems, employing matrix multiplication for computations. Convolutional Neural Networks leverage an algorithm designed for Computer Vision applications, with layers organized in three dimensions to enhance spatial awareness. Recurrent Neural Networks excel in tasks requiring memory of past decisions, although their extensive use of Linear Algebra concepts poses challenges for hardware implementations. Due to the complex self-attention mechanisms and the large number of parameters that need iterative optimization, the training of Transformer Neural Networks (TNNs) demands a substantial amount of computational resources and time. Spiking Neural Networks (SNNs) exhibit great potential for neuromorphic computing, closely mimicking human brain functionalities. However, these networks rely on the leaky integrate-and-fire approach, which introduces integration complexities and may impede efficiency.
\par
While RNNs, TNNs, and SNNs demonstrate considerable promise in deep learning applications, their limited utilization is attributed to the constraints and efficiency barriers they impose on digital AI hardware. Realizing their full potential calls for innovations in AI hardware, enabling their integration to provide substantial benefits without undesirable consequences. In this pursuit, we look to the past, exploring the possibility of analog technologies, potentially integrated with digital hardware, as the key to unlock next-generation AI advancements.

\subsection{Analog Deep Learning}
Analog deep learning refers to the utilization of analog computing principles and hardware to perform the computations involved in deep learning tasks. This approach leverages the continuous nature of analog signals to potentially achieve higher energy efficiency and faster processing compared to traditional digital methods, which rely on discrete representations\cite{hu2018memristor}.

Digital Deep Learning has garnered praise for its accuracy and ease of implementation, exploring the analog counterpart could address several of its limitations, such as efficiency and further advancement. Analog devices, on the other hand, offer diverse physical properties that can be harnessed to achieve varied results, akin to leveraging different neural network architectures for AI applications. Prominent examples of analog properties include Cation Migration \cite{cation1,cation2, cation3,cation4,cation5,cation6,cation7}, Anion Migration \cite{anion1,anion2,anion3,anion4,anion5,anion6}, Ferroelectric Gating \cite{ferroelectric1,ferroelectric2,ferroelectric3,ferroelectric4,ferroelectric5,ferroelectric6,ferroelectric7,ferroelectric8}, Phase Change Systems \cite{pc1,pc2,pc3,pc4,pc5,pc6,pc7}, Metal-to-Insulator Transition \cite{mit1,mit2}, Superconductivity \cite{sc1,sc2,sc3,sc4,sc5,sc6,sc7}, and Spintronics \cite{spin1,spin2,spin3,spin4,spin5,spin6,spin7}.
\par
Analog devices, like resistive crossbar arrays \cite{onen2022nanosecond, bigReview}, have gained popularity as alternatives to GPU-driven ANNs due to their inherent simplicity. Instead of relying on digital matrix multiplication, which becomes increasingly taxing as the number of neurons increases, analog hardware leverages the calculations themselves using analog components. Resistive crossbar arrays exemplify this approach. By arranging resistors with conductance values corresponding to the network's weights in an array, connected to other resistors via conductive elements, direct modeling of matrix multiplication occurs within the hardware. Applying a chosen voltage through the device results in currents through specific resistors, determined by Ohm's Law. By utilizing Kirchhoff's Junction Rule, the currents from individual resistors can be summed together via conductive connections to a main output wire.
\par
With various analog implementation methods available, we can explore and identify the optimal solution for specific AI tasks. However, analog devices exhibit greater sensitivity to system noise compared to their digital counterparts. Technologies like Cation Migration, Anion Migration, and Spintronics showcase high analog characteristics, leading to significant variation under different operational conditions. As a result, controlling these technologies can prove challenging. Yet, by employing low-noise design and integrating them with digital circuits, analog devices may become more controllable than ever before, presenting exciting possibilities for Analog Deep Learning.

\subsection{Neuromorphic Computing}
Neuromorphic computing \cite{bigReview} is an approach to designing computer systems and hardware that draws inspiration from the structure and functioning of the human brain's neural networks. It aims to mimic the brain's ability to process information using highly interconnected neurons and synapses, enabling efficient and parallel processing of complex tasks. Neuromorphic computing systems are designed to perform tasks like pattern recognition, sensory processing, and decision-making with lower power consumption and potentially higher efficiency compared to traditional computing architectures. These systems leverage analog components and specialized hardware to emulate neural behavior and achieve biologically inspired computational capabilities. 

In the context of analog deep learning, neuromorphic computing refers to an innovative approach that capitalizes on the principles of the human brain's neural networks. It intricately combines principles from both neuroscience and analog electronics. In other words, neuromorphic computing systems are designed to replicate the brain's intricate connectivity and information processing mechanisms using analog components. This enables efficient and parallel computation of complex tasks, such as pattern recognition and decision-making, while potentially offering advantages in energy efficiency compared to traditional digital architectures. By simulating neural behavior through specialized hardware, neuromorphic computing aligns with the goals of analog deep learning by leveraging continuous signals and emulating neural computation for enhanced performance in certain machine learning applications.

\section{Taxonomy and Description of Analog Implementations}

We classify the current state of Analog Deep Learning (ADL) implementation techniques into four main types: (a) Ion Migration, (b) Charge Distribution, (c) Material-Level Changes, and (d) Magnetization. 



\begin{figure*}
    \centering
    
    \includegraphics[width=1.5\columnwidth]{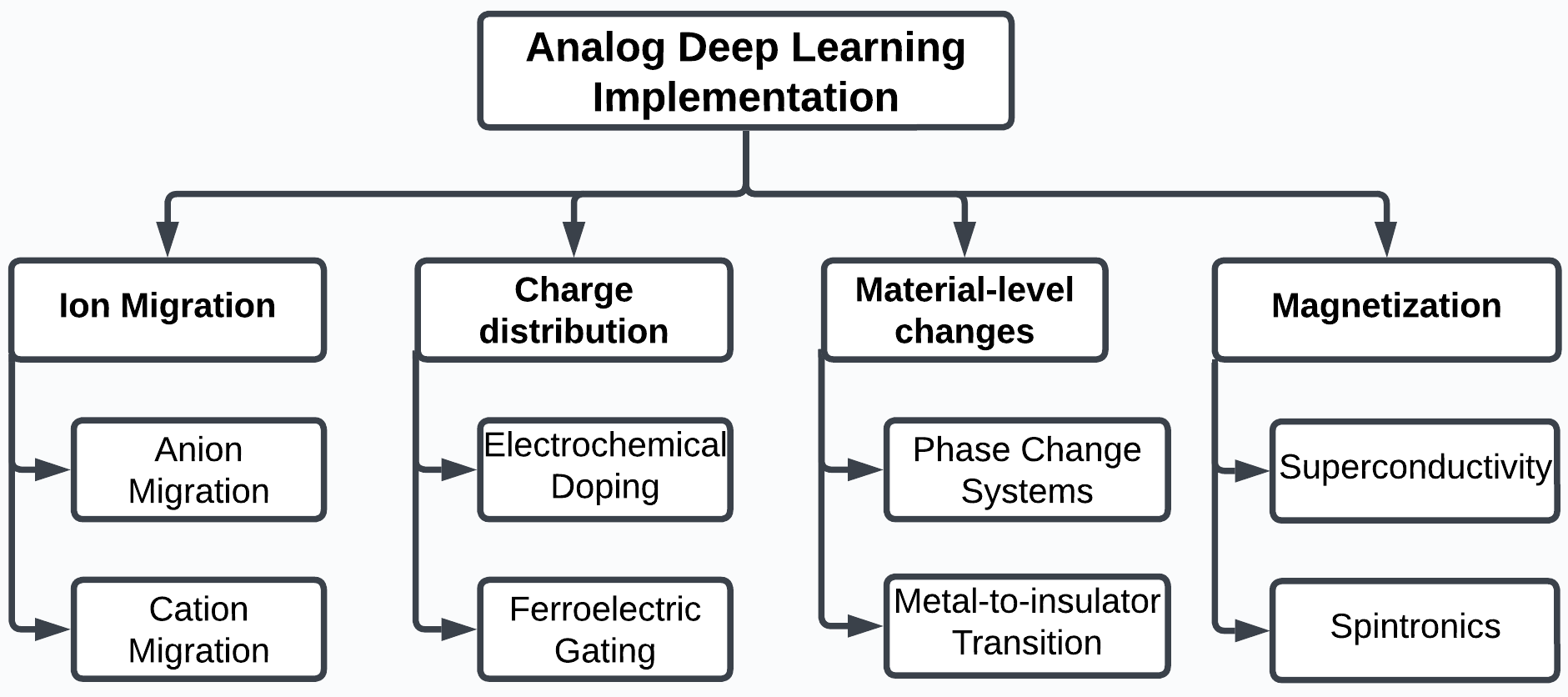}
    \caption{Taxonomy of Analog Deep Learning}
    \label{figure1}
\end{figure*}

\subsection{Ion Migration}
Ion migration has demonstrated significant promise in accomplishing a wide range of crucial neural functions that replicate biological systems. The core concept behind ion migration revolves around the careful manipulation of ionic movements within a material. This manipulation is typically achieved through external stimuli, such as optical pulses \cite{wang2020optically, zhou2019optoelectronic, park2020retina, ahmed2021fully} or electric fields \cite{onen2022nanosecond, van2017non, nguyen2022dual, wang2018fully, kim2022vertically, kwak2021halide}. The primary goal of this process is to induce changes in the material's inherent properties, which are readily measurable. For instance, one of the most notable changes is in the material's conductance.
Materials that facilitate the movement of cations or anions offer a promising avenue to achieve electricity conduction beyond traditional metals. When bias voltages are applied, mobile cations or anions can be attracted towards the bias direction, effectively promoting charge flow within the material. This dynamic mechanism presents significant possibilities for manipulating electric conductivity in a controlled manner, enabling applications for neuromorphic computing.

It is to be noted that the process of ion migration is not just about moving ions, rather, about precisely controlling how these ions move within a material's structure in response to external influences. By applying stimuli like light (optical pulses) or electricity (electric fields) as mentioned before, the ions within the material are set in motion in a controlled manner. This movement leads to alterations in the material's fundamental characteristics. Conductance, a property that measures how easily electricity can pass through a material, is a key attribute that can be effectively modified through this process. Such modifications are crucial as they are the basis for the material's ability to mimic neural functions. When delving deeper into the mechanisms of ion migration, we can categorize the various implementations of this principle based on the type of ion movement they utilize. These categories include cation filament-based approaches and anion migration-based methods. Each of these categories represents a different strategy in harnessing ion migration to achieve the desired changes in material properties.

In cation filament-based methods, the focus is on the movement of positively charged ions (cations). This movement creates conductive pathways, or 'filaments', within the material. These filaments play a significant role in altering the material's conductance and are a key aspect of how these devices can mimic neural functions.
On the other hand, anion migration-based approaches involve the movement of negatively charged ions (anions). This type of ion migration also leads to changes in the material's properties, offering a different mechanism to achieve neuromorphic functionalities.

The study and application of ion migration in neuromorphic devices are at the forefront of merging material science with neural emulation. This convergence is critical for advancing computing technologies and artificial intelligence, pushing the boundaries of what can be achieved with traditional computing paradigms. Ion migration is fundamentally important in analog deep learning for several reasons:

\begin{enumerate}

\item \textbf{Emulation of Biological Neural Functions:} One of the primary goals of artificial neural networks is to mimic the way biological brains process information. Ion migration in neuromorphic devices allows for the emulation of a wide array of neural functions that are biomimetic, implying that they closely replicate biological processes. This is crucial for developing computing systems that can mimic the efficiency and adaptability of the human brain.

\item \textbf{Controlled Manipulation of Material Properties:} The ability to precisely control ionic movements within a material in response to external stimuli, such as optical pulses or electric fields, is the main principle of ion migration. This control is key to inducing changes in the material's inherent properties, such as conductance, which is essential for neuromorphic computing. The conductance changes in materials can be used to simulate the synaptic plasticity found in biological neurons.

\item \textbf{Versatility in Implementation:} Ion migration can be implemented in various ways, such as through cation filament-based methods or anion migration-based approaches. Each method offers a unique strategy for altering material properties, providing flexibility in designing neuromorphic systems. This versatility is important for tailoring neuromorphic devices to specific computational tasks or to replicate particular neural functions.

\item \textbf{Creation of Conductive Pathways:} In cation filament-based methods, the movement of positively charged ions creates conductive 'filaments' within the material. These filaments are crucial for altering the material's conductance, akin to how neural pathways work in the brain. This aspect is vital for creating neuromorphic devices that can dynamically change their connectivity patterns, similar to learning and memory formation in biological systems.

\end{enumerate}

\subsection{Semiconductor-based Charge Distribution}
Among the diverse array of materials that facilitate the movement of cations or anions, semiconductors hold a unique position, blurring the lines between conductors and insulators. Semiconductors possess the ability to function as both conductors and insulators, allowing for analog-like modulation of their conductance. This exceptional property enables us to manipulate semiconductors' conductivity using bias voltages, achieved through Ferroelectric Gating, or by introducing charge imbalances via Electrochemical Doping, thereby making it suitable for neuromorphic computing.

The key to a semiconductor's behavior lies in its energy band structure. It has a valence band (filled with electrons) and a conduction band (where electrons can move freely). The gap between these bands (bandgap) is small enough that electrons can move from the valence to the conduction band under certain conditions, like the application of heat or light. This ability to transition between conductive and insulative states is what makes semiconductors so valuable. They can be manipulated to either allow or prevent the flow of electricity. Unlike digital components that operate in binary (on/off), semiconductors can exhibit analog-like behavior. This implies that their conductance can be modulated in a continuous range, similar to how the human brain modulates the strength of synaptic connections. There are two main types of semiconductor-based mechanisms for implementation of analog deep learning hardware - (a) Ferroelectric Gating and (b) Electrochemical Doping. Ferroelectric materials have an electric polarization that can be reversed by the application of an external electric field. By using ferroelectric materials as a gate in semiconductor devices, the conductivity of the semiconductor can be modulated. The ferroelectric gate can induce or remove electric fields, effectively controlling the movement of electrons in the semiconductor. On the other hand, Electrochemical Doping involves introducing impurities into a semiconductor to change its electrical properties. Ions are introduced into the semiconductor, creating charge imbalances. This process alters the semiconductor's electrical characteristics, particularly its conductance. The movement of cations or anions in this process is crucial. Depending on the type of ion and its concentration, the semiconductor's ability to conduct electricity can be significantly altered. Their unique position between conductors and insulators and their ability to modulate conductance are pivotal in the development of neuromorphic computing systems. Their capacity to emulate neural functions through techniques like Ferroelectric Gating and Electrochemical Doping makes them ideal for creating computing systems for deep learning that closely resemble the efficiency and adaptability of biological brains. These two processes are highly relevant to Analog Deep Learning in the following ways:

\begin{enumerate}

\item \textbf{Mimicking Neural Functions:} The brain's neural networks modulate synaptic strengths in a continuous, analog-like manner. Semiconductors, with their ability to have finely tuned conductance, can mimic this aspect of neural behavior.

\item \textbf{Dynamic Adaptability:} Just as biological neurons adjust their synaptic strengths based on stimuli, semiconductors can dynamically alter their conductive states in response to external inputs like electric fields or ion concentrations.

\item \textbf{Energy Efficiency:} Semiconductors can operate at lower power levels compared to traditional digital circuits, which is a key consideration in mimicking the energy-efficient nature of the human brain. This is also a huge advantage over the conventional digital deep learning pathway.

\item \textbf{Complex Computations:} The ability to modulate conductance in a continuous range allows for complex, non-linear computations, which are essential for tasks like pattern recognition and decision-making, mirroring brain-like processing.
    
\end{enumerate}

\subsection{Material-Level Changes}
The properties at the material-level play a crucial role in determining the conductance of a material. Characteristics such as temperature and surface area have been observed to have a significant impact on the conductance behavior of different materials. By altering these material-level characteristics, we can effectively modulate the conductance of the material. Notably, controlling the temperature or adjusting the surface area can lead to effective transitions, converting a material from a metal to an insulator or vice versa, making it highly applicable to neuromorphic computing.

In materials, the movement of electrons, which carry electrical current, is affected by temperature. As temperature changes, it alters the energy levels of electrons, impacting their ability to move and thus the material's conductance. At higher temperatures, electrons gain more energy, which can increase their mobility in some materials, leading to higher conductance. Conversely, in other materials, increased thermal energy can lead to more electron scattering, reducing conductance. Certain materials exhibit phase transitions at specific temperatures where they change from conductors to insulators or vice versa. For example, some materials become superconductors at very low temperatures, where they conduct electricity with zero resistance.

The surface area of a material affects the pathways available for electron movement. In materials with a larger surface area, there are more pathways for electrons to travel, which can influence the material's conductance. In nanostructured materials, where the surface area-to-volume ratio is high, surface effects become more pronounced. This can lead to unique conductive properties that differ significantly from bulk materials. The surface of a material can have different electronic states compared to its bulk. Additionally, surfaces may participate in chemical reactions that can alter the material's conductive properties.

This capability to manipulate the the conductance of some materials is key to developing neuromorphic computing systems that can emulate the complex and energy-efficient nature of biological neural networks, discussed as follows:

\begin{enumerate}

\item \textbf{Material-Level Control:} As discussed earlier, by controlling temperature and surface area, it is possible to fine-tune the conductance of materials. This control is essential for creating neuromorphic devices that can mimic the variable synaptic strengths found in biological neural networks. These synaptic strengths are commonly called connection weights in artificial neural network which are its learnable parameters. This material-level control allows for the creation of physical synapses whose strengths can be adjusted, akin to the weight adjustments in digital neural networks.

\item \textbf{Metal-Insulator Transitions:} The ability to transition a material from a metal (conductive) to an insulator (non-conductive) or vice versa is particularly intriguing. This property can be harnessed to create switches and memory elements in neuromorphic circuits, similar to how neurons and synapses function in the brain.

\item \textbf{Application in Synaptic Emulation:} In analog neural computing, materials that can undergo such transitions are valuable for emulating the synaptic plasticity of neurons – the ability of synapses to strengthen or weaken over time, which is fundamental to learning and memory.

\item \textbf{Energy Efficiency and Complexity:} By leveraging materials whose conductance can be precisely modulated, it is possible to create more energy-efficient and complex computational architectures. This is because analog circuits can be more energy-efficient than their digital counterparts, especially when they operate using materials with tunable conductance. By precisely controlling material properties, these circuits can perform computations with less power, which is a significant advantage for deep learning applications that require extensive computation.

\item \textbf{Implementing Non-Linear Activation Functions:} Deep learning involves non-linear transformations, typically through activation functions like ReLU or sigmoid. Analog hardware can implement these functions physically using materials whose conductance properties change non-linearly with temperature or surface area.

\end{enumerate}

\subsection{Magnetization}

According to Electromagnetic Theory, the arrangement of a conductor in a loop induces a magnetic field around it. This magnetic field's strength and distribution are intrinsically linked to the conductance of the material. By manipulating external influences on the magnetic field, one can effectively control the conductance of the material. However, beyond the influence of external magnetic fields, the subatomic level presents another avenue for manipulating conductance. A material's conductance can be affected by the alignment of electron spins within it. Analogously changing the alignment of electron spins across a device provides a means to finely control its electrical conductance. Again, manipulation of conductance can be leveraged to mimic varying synaptic strength of neural networks. 

The ability to control the conductance of materials through magnetic fields and electron spin alignment presents a novel and potentially powerful approach for developing analog deep learning hardware. This approach can lead to more energy-efficient, adaptable, and biologically realistic computing systems, pushing the boundaries of what is currently possible in artificial intelligence and machine learning, as follows:

\begin{enumerate}

\item \textbf{Simulating Synaptic Weights:} In neural networks, the strength of connections (synaptic weights) between neurons is fundamental to learning. By using materials whose conductance can be controlled via magnetic fields and electron spin alignment, analog deep learning systems can simulate these synaptic weights. The conductance can be finely tuned to represent different weight values.

\item \textbf{Energy-Efficient Computation and scalability:} This method of controlling conductance can be more energy-efficient compared to traditional voltage-based methods. Lower energy consumption is a significant advantage for deep learning systems, which typically require extensive computational resources. Techniques based on magnetic fields and electron spins might offer pathways to more scalable and miniaturized designs, crucial for developing compact yet powerful deep learning systems.

\item \textbf{Implementing Non-Linear Dynamics:} The relationship between magnetic fields, electron spin alignment, and conductance can introduce non-linear dynamics into the system. This is beneficial for deep learning, as non-linear transformations are essential for complex pattern recognition and decision-making tasks.

\item \textbf{Dynamic Adaptability:} Just as synaptic weights in biological brains change in response to stimuli, the conductance in these systems can be dynamically adjusted in response to changes in magnetic fields or electron spin configurations. This allows for on-the-fly learning and adaptability. These systems can process information in an analog manner, dealing with continuous values rather than binary, which is more akin to how biological neural networks operate and can motivate a new direction in artificial neural network tailored to the strengths of magnetic field and spin-based conductance control.

\end{enumerate}

\section{Analog Deep Learning Performance}
\subsection{Algorithms, Applications and Accuracy}
While digital systems are praised for their accuracy, their analog counterparts are met with higher levels of noise due to factors such as Brownian motion and heat distributions. Neural networks rely on the steady state of its weights, often represented by the conductance or resistance values of components in analog crossbar arrays \cite{pc3, onen2022nanosecond, bigReview}. With random noise interference in the system, the weights may easily change their values and heavily impact the functionality of the network itself. To address this issue, efforts should be focused on making design decisions for analog components that minimize the amount of noise in a device. This process begins with considering the physical properties utilized by the device. In the following subsections, we briefly discuss the key working mechanisms, application areas, and performance achieved by different state-of-the-art analog deep learning methods.
\\

\subsubsection{Cation Migration}
One of the most versatile approaches to analog deep learning computing devices is Cation Migration \cite{cation1, cation2, cation3, cation4, cation5, cation6,cation7}, which involves the oxidation and reduction of a material's metal ions and the induced movement of cations to control the synpatic weights in a neural network. Most neuromorphic devices that utilize Cation Migration use memristive crossbar arrays to directly realize neural networks, and employ resistive switching for each memristor. Applying a carefully controlled voltage across an electrolytic layer can allow for easy and stable conductance modulation for synaptic transistors that can be placed in a crossbar array\cite{cation7}. The material of a memristor, a type of synaptic transistor, can heavily impact its performance in a neural network. We discuss the main cation migration-based deep learning attempts below:
\\

\noindent
\textbf{(a) Perovskite memristors}: A recent study showed that perovskite memristors can act as drift-based non-volatile weights, allowing them to be stable memory components\cite{cation3}. Due to its electrically-controllable nature, Cation Migration was able to minimize the error involved in inherently analog operations. The study involved testing a network on the classification of four common neural firing patterns observed in the human brain: Bursting, Adaptation, Tonic, and Irregular. The network was part of a fully-memristive reservoir computing (RC) framework, interfaced with an artificial neural network (ANN), and demonstrate the advantages of reconfigurable memristive materials. The reservoir was modeled as a network of recurrently-connected units, functioning as short-term memory. Temporal signals entering the reservoir underwent high-dimensional nonlinear transformations, enhancing the separability of their temporal features. A Linear Classifier Layer was connected to the reservoir units with all-to-all connections with 30 inputs, 4 outputs, and one bias unit. The network was trained with a batch size of 1 and suitably tuned hyperparameters to perform classification based on the temporal information stored in the reservoir. The work particularly highlighted the advantages of perovskite memristors in a fully-memristive RC framework. The framework included a dynamically-configured layer of virtual volatile reservoir nodes and a readout ANN layer with non-volatile weights. The reconfigurability addressed three computational requirements: accumulating/decaying short-term memory, stable long-term memory for retaining trained weights, and a circuit methodology for accessing analog states from binary devices.

Another similar work \cite{cation4} used perovskite nanocrystal memristors to simulate a three-layered artificial neural network (ANN) in image recognition tasks. Two datasets were employed for training the ANN: a small image version (8 × 8 pixels) of handwritten digits from the “Optical Recognition of Handwritten Digits” dataset and a large image version (28 × 28 pixels) of handwritten digits from MNIST dataset. The network was trained for 25 epochs using backpropagation and it achieved accuracies: $94\%$ for small images and $88\%$ for large images. The recognition accuracy was observed to be comparable to that of $\alpha$-MoO3 based synaptic devices. 
\\

\noindent
\textbf{(b) Tin Sulfide memristors:} SnS (Tin Sulfide) memristors \cite{cation1} were used for simulating a Recurrent Neural Network (RNN) that achieved a stable accuracy level of $85\%$ in classifying Korean sentences with standard deviations up to $\sigma = 0.6 ~nA$ across different models. The authors implemented high-dimensional, nonlinear reservoir computing (RC) using two-dimensional memristors based on tin sulfide (SnS). These memristors are unique due to their dual-type defect states associated with Sn and S vacancies. The vacancies of Sn, S, and or both atoms in the makeup of some memristors can keep the conductance modulation somewhat consistent since the same modifications are made to memristors throughout the network. The classification was achieved based on the current flowing through five memristors in the reservoir at the final time step, which were used as inputs to the readout layer. Various supervised learning models like single-layer perceptron, support vector machine (SVM), and logistic regression were also implemented. This study showcased the integration of advanced machine learning techniques with novel hardware (memristors) for efficient language processing tasks.
\\

\noindent
\textbf{(c) Cesium Halide Memristors:} Su et al. \cite{cation2} trained a three-layer fully connected neural network using Cesium Halide Memristors (CHM) via a back-propagation algorithm. Two datasets were tested: an 8×8 pixel image version of handwritten digits from the Optical Recognition of Handwritten Digits dataset, and a 28×28 image version MNIST database. The simulation accounted for real-world imperfections like conductance variations during weight updates and nonlinearity. The classification accuracies for both datasets were recorded over each training epoch. For the smaller 8×8 handwritten digits dataset, the accuracies were higher than $89\%$ for CsI and $90\%$ for CsBr, deviating only by $6\%$ from the ideal value simulated with 32-bit floating-point numeric precision. In the case of the MNIST dataset, accuracies exceeded $85\%$ in the second epoch for both devices, reaching $89.4\%$ for CsI and $88.4\%$ for CsBr after 40 epochs. 
These results were compared with other material systems like TaOX that achieved $80\%$ accuracy and GST phase-change memory that achieved $82.2\%$ accuracy. The CHMs (CsI and CsBr) used in this study showed significant improvements in nonlinearity, leading to much higher recognition rates. This demonstrates the potential of using CHMs as basic computing elements in analog deep learning systems, highlighting their superiority in terms of recognition accuracy compared to other materials.
\\

\noindent
\textbf{(d) pV3D3 memristor:} A pV3D3 memristor-based flexible synapse device \cite{cation6} has been used to design an ANN with a crossbar architecture that consisted of 32×32 input neurons, 10 middle neurons, and 3 output neurons. It was trained for a facial recognition task using nine 32×32 pixel training images extracted from the Yale Face Database. After 24 epochs of training, the ANN, based on the flexible pV3D3 memristor, achieved an average recognition rate of $88\%$, misclassifying 3 out of 24 images. This performance is slightly lower but comparable to Yao et al.'s study, which achieved a $91\%$ recognition rate using a one-layer perceptron with a 1T-1R memristor array. Additionally, the recognition rate for the MNIST handwritten digit dataset was approximately $86\%$ with the pV3D3 memristor, surpassing the performance of a carbon nanotube synaptic device-based spiking neural network.
\\

\subsubsection{Anion Migration}
Anion migration leads to alterations in material properties, such as conductivity, caused by the targeted movement of anions, including oxygen ions, sulfur ions, and organic anions. Devices that use Anion Migration are very similar to those who use Cation Migration in that both involve inducing the movement of ions to modulate synaptic weights, but the former involves the movement of anions rather than cations. Anions induced to move by a bias voltage can cause changes in a material's conductivity. If the resistive element is a filament, then these changes are amplified enough that small changes in the bias voltage can significantly change the conductance of the filament\cite{bigReview}. Anion Migration, coupled with memristors in a crossbar array, has proven to be very efficient because of the simplicity and lack of extra regulators, sensors, and heavy conversion processes\cite{anion4}.
\par
Since a material's negative charge carriers tend to be larger and more controllable, Anion migration can be an excellent way to keep a neuromorphic device's systems under control. Anion Migration devices don't fall short on accuracy either, as memristors employing Anion Migration in a crossbar array had an accuracy of 96\% when tested with MNIST image classification\cite{anion5}. Ion migration techniques, both cationic and anionic, are relatively easy to implement and test compared to other techniques becuase of easy ability to control bias voltages, both manually and through automation.
\\

\noindent
\textbf{(a) Metal oxide memristors:} 
Prezioso et al.\cite{prezioso2015training} used transistor-free metal-oxide memristor crossbars to train a single-layer perceptron using a coarse-grain version of the delta rule algorithm to accurately classify 3x3-pixel black/white images into three distinct categories (symbolizing letters). The authors reported perfect classification results after 23 epochs. A similar work by Yan et al. \cite{yan2021reconfigurable} introduced tin oxide-based memristors for implementing and optimizing Boltzmann Machine (BM). They demonstrated that reconfigurable tin oxide (SnOx)/molybdenum disulfide (MoS2) heterogeneous memristive device is capable of producing exponential-class stochastic sampling. This device featured dynamically adjustable distribution parameters, facilitating the hardware implementation of the Boltzmann Machine (BM) that requires temperature regulation. Consequently, this allowed for the precise implementation of the necessary cooling strategy. The BM prototype comprised six stochastic units, each equipped with a tin oxide/MoS2 heteromemristor that exhibited a roughly sigmoidal switching probability in response to applied voltages, along with accompanying peripheral circuitry.
Another Au/Ni/HfO2/Ni/SiO2 memristor was developed by Hu et al. \cite{hu2015associative} for the simulation and construction of a Memristive Hopfield Neural Network (MHN) comprising 6,561 synapses, equivalent to an 81x81 matrix. The authors successfully developed a 3-bit MHN that had nine synapses with 6 memristors and 3 neurons. The synaptic weights in the MHN were programmable, allowing for easy adjustment to positive or negative values by altering the conductance of the memristors. 
\\

\noindent
\textbf{(b) Semiconductor alloy random access memory:} Another line of work used single-crystalline Silicon Germanium (SiGe) epitaxial random access memory (epiRAM) for simulation of an artificial neural network designed for supervised learning using the MNIST handwritten recognition dataset. The network comprised a three-layer structure with 28×28 neurons in first layer, 300 hidden neurons, and 10 output neurons, employing a multilayer perception (MLP) algorithm with stochastic gradient descent for weight updates. This simulation incorporated non-ideal factors of epiRAM properties, including finite on/off ratio, limited conductance levels, device-to-device variation, cycle-to-cycle variation, wire resistance, and read noise. The 784 input layer neurons corresponded to the pixels of a 28×28 MNIST image, while the 10 output layer neurons represented the 10 digit classes (0–9). The process involved the inner product of the input neuron signal vector and the first synapse layer, followed by activation and binarization, which then served as the input for the second synapse array. The synapse layer consisted of epiRAM crossbar arrays and peripheral circuits. The network adjusted synapse weights based on the delta weight calculated from the inner product of the input signal vector and the synapse vector. After training with one million patterns randomly selected from 60,000 images in the training set, the network's recognition accuracy was tested with 10,000 images from a separate testing set. The simulation demonstrated that the epiRAM-based neural network could achieve an average recognition accuracy of $95.1\%$.
\\

\noindent
\textbf{(c) Optically driven semiconductors:} Ahmed et al. \cite{ahmed2021fully} introduced a fully light-modulated  semiconductor device and applied it for the simulation of Optical Neural Networks (ONNs). The networks were trained for the classification of 28 × 28 pixel images of handwritten digits from the MNIST dataset using a single-layer perceptron model. This model also employed supervised learning with a back-propagation algorithm. It consisted of 784 input neurons corresponding to the image pixels and 10 output neurons. The input neurons received pixel values, which were transformed into output values through a weight matrix and then processed by a sigmoid activation function. The synaptic weights in the network were updated based on the difference between the output value and the label value, using a backpropagation algorithm. The synaptic weight calculation generally considers both positive and negative values, while the conductance of the BP devices used is always positive. To address this, synaptic weight was treated as the difference between two conductance values, derived from experimental data. The recognition accuracy was evaluated for different pulse widths of optical stimuli used in the BP devices. The ONN achieved a maximum recognition accuracy of approximately $90\%$ with a 1 ms pulse width, surpassing previously reported neural networks with similar topology. The study also highlighted the stability of the network over extended training cycles, demonstrating its potential in neuromorphic computing applications.

Zhou et al. \cite{zhou2019optoelectronic} developed two-terminal optoelectronic resistive random access memory (ORRAM) synaptic devices with a Pd/MoOx/ITO (indium tin oxide) structure. It was capable of ultraviolet (UV) light sensing, optically triggered non-volatile and volatile resistance switching, and light-tunable synaptic behaviors. For testing, an image database containing 6×7 pixel images of the letters P, U, and C was used. The study employed a three-layered artificial neural network with input, hidden, and output nodes being 42, 20, and 3 respectively. It was trained through a back-propagation (BP) algorithm with a fixed learning rate of 0.01 and sigmoid activation functions in the hidden and output layers. The ORRAMs were directly connected to the input layer of the neural network, and images were processed through the ORRAM or delivered to the input layer individually during training. Remarkably, the inclusion of the ORRAM array for preprocessing significantly reduced the number of epochs required to achieve a $99\%$ recognition accuracy—from 3740 epochs without ORRAM to 2186 epochs with ORRAM, a $41.5\%$ reduction. 
\\

\noindent
\textbf{(d) Transistor-memristor arrays:}
A key computational function involving memristors is vector matrix multiplication (VMM), which can be executed in compact crossbar layouts through a single analog computation step. This process leverages Ohm’s law for electrical resistance and Kirchhoff’s current law for current summation. Hu et al. \cite{hu2018memristor} proposed effective VMM via integrated transistor-memristor arrays and used it for training and implementation of a single-layer neural network towards handwriting recognition using the MNIST dataset. The network was trained in MATLAB using softmax regression. For classification, outputs were exponentiated and normalized to create a probability distribution, a step not implemented in the Dot Product Engine (DPE) due to the inability of current memristor crossbars to compute the nonlinear function directly. A software-trained network was programmed into a 96 × 40 portion of a 128 × 64 memristor crossbar, representing the largest demonstration using memristor crossbars in hardware to date. To fit the network into the DPE platform, the MNIST images were resized to 19 × 20, and the network was retrained. The network, which classified 784 pixel inputs into 10 classes, was reshaped into a 96 × 40 array without degrading recognition accuracy, which remained at $92.4\%$. For digit classification using the DPE, input images were processed and partitioned, with voltage signals applied to the rows of the memristor array programmed to the trained weight matrix. With linear correction, the hardware DPE neural network implementation reached an accuracy of $89.9\%$, only a $2.5\%$ reduction compared to the ideal software accuracy, with the remaining loss attributed to device programming inaccuracies and cumulative wire resistances.

All these works demonstrated two-layer and three-layer memristor multi-layer perceptrons for image recognition using the MNIST database. However, a complete Convolutional Neural Network (CNN), crucial for more complex image recognition tasks, were not fully realized in a memristor-based hardware system. This gap was primarily due to several challenges: poor yield and non-uniformity of memristor crossbar arrays, difficulty in matching software-level performance due to device imperfections like variations, conductance drift, and device state locking, and the time-consuming nature of the convolutional operation in CNNs, which is typically sequential and leads to speed mismatches in the hardware. Addressing these challenges, a versatile memristor-based computing architecture \cite{yao2020fully} was proposed for neural networks, featuring a memristor cell with a TiN/TaOx/HfOx/TiN material stack and continuous conductance-tuning capability. This architecture enabled the successful demonstration of a complete five-layer memristor-based CNN (mCNN) for MNIST digit image recognition. The optimized material stacks facilitated reliable and uniform analog switching behaviors in 2,048 one-transistor–one-memristor (1T1R) arrays. Employing a hybrid-training scheme, the experimental recognition accuracy of this system reached $96.19\%$ for the entire test dataset, marking a significant advancement in memristor-based neural network hardware.
\\

\noindent
\subsubsection{Electrochemical Doping}
Electrochemical (EC) Doping, as applied to Analog Deep Learning, involves modulating the physical properties of the device's material to alter the electrolyte layers and thus change the device's conductance. Due to the nature of doping in conductive materials, EC Doping can provide greater accuracy in neural networks and is much less prone to noise than the other methods. A bias voltage is used to control the flow of particles inside the material. EC Doping can allow memristors to be in a variety of non-volatile states, as one study found a value retention of about 25 hours for an electrochemically-doped thin-film transistor\cite{ec2}, thereby increasing likelihood of an accurate inference by the neural network. Another study found that Dual-mode Organic Electrochemical Transistors (OECTs) can operate in both depletion and enhancement modes using a self-doped conjugated polyelectrolyte, with changes in the bias voltage also switching modes\cite{ec1}. However, its application has been limited to basic computational operations such as logic gates, and it has not yet been utilized in neural networks, despite their significant potential. Only one type of electrochemically-doped transistor has been successfully demonstrated to train neural networks, as discussed below.
\\

\noindent
\textbf{ENODe:}
A novel electrochemical neuromorphic organic device (ENODe), fundamentally different from  memristors,\cite{cation5} was recently proposed by Burgt et al. The authors simulated a 3-layered neural network, which they trained and tested on three distinct datasets: an 8 × 8 pixel image version of handwritten digits, the MNIST dataset comprising 28 × 28 pixel images of handwritten digits, and a Sandia file classification dataset. The training of this neural network resulted in an accuracy ranging between $93\%$ and $97\%$ that was higher than the accuracy achieved by memristor-based networks. Notably, this performance was consistently within $2\%$ of the ideal performance achievable by a floating-point-based neural network, which represents the theoretical limit for this algorithm. This close proximity to the theoretical limit underscores the effectiveness of ENODes  in handling diverse datasets.
\\

\noindent
\subsubsection{Ferroelectric Gating}
The Ferroelectric Field-Effect Transistor (FeFET) has gained many improvements over the past years, especially towards neuromorphic computing. This technology employs ferroelectric polarization, which alters the movement of particles in a material, thus changing its conductance.
\par
However, there are many variations of the technology such as the use of optical modulation and domain-wall switching, which one study found could be a simpler, more efficient way for ferroelectric switching\cite{ferroelectric3}. There are many other benefits of ferroelectric gating, such as the detection of small variations in analog signals\cite{ferroelectric1} and a closer mimicry of biological neurons using Hafnium Oxide-based FeFETs\cite{ferroelectric7}.
\par
Ferroelectric gating is very similar to semiconductor doping for transistors in that both change the electrical properties of materials, but allows for more steady-states for memristors than doping does for transistors.
\\

\noindent
\textbf{(a) Ferroelectric Field-Effect Transistors (FeFETs):}
Gao et al.\cite{ferroelectric6} significantly advanced the field by incorporating interfacial states into a ferroelectric Hf0.5Zr0.5O2 thin film through a carefully controlled annealing process. This approach led to the creation of a multifunctional two-dimensional Ferroelectric Field-Effect Transistor (Fe-FET). The device harnesses the synergistic effects arising from ferroelectric polarization and charge trapping behavior, resulting in a single device with a diverse range of capabilities.
\par
Additionally, this multifunctional device exhibits reliable memory properties, allowing for the storage of information at multiple levels with extended retention times. Beyond traditional memory functions, the Fe-FET showcases tunable synaptic behaviors, offering adaptable plasticity that encompasses both short-term plasticity (STP) and long-term plasticity (LTP). This adaptability mimics the flexible nature of biological synapses.
\par
Moreover, the device demonstrates well-developed photodetection capabilities, showcasing its versatility in different applications. The integration of ferroelectric polarization and charge trapping not only enhances memory retention but also contributes to the emulation of synaptic behaviors. In simulations of artificial neural networks, the Fe-FET achieves a pattern recognition accuracy of 81\%, underscoring its efficacy in complex cognitive tasks.
\\

\noindent
\textbf{(b) Ferroelectric Tunnel Field-Effect Transistors (FeTFETs):}
Within the broader category of FeFETs, FeTFETs use gate voltages to tune a tunneling barrier through which electrons can flow. When the FeTFET is in the ON state, the barrier is shrunk by a high voltage which propels electrons across the barrier, and a low voltage widens the barrier in the OFF state and thus impedes electron flow. Using this technology, Zhu et al.\cite{ferroelectric1} explored the detection of small variations in analog signals. The researchers investigate the detectability of individual defects in the gate oxide, particularly focusing on the impact of these defects on the transfer characteristic in FeTFETs. The study reveals that individual defects in the oxide lead to fluctuations in current (ID) during the sweeping of gate bias voltage (VGS), both before and after ferroelectric switching. The most significant current peaks with the largest fluctuations are examined to demonstrate the effect of individual defects. The fluctuations are attributed to the capture and emission of electrons by localized, positively charged defects. The presence of these defects is observed through random telegraph signal (RTS) noise, and the time constants for electron capture and emission are determined. Before ferroelectric switching, a two-level RTS noise is observed, indicating one defect trap. After switching, a three-level RTS noise is observed, suggesting the presence of two traps. The analysis of the higher current levels (trap 2) after switching reveals different time constants for electron capture and emission. The study also estimates the depth (z) of the defect from the channel into the oxide, suggesting that the probed individual defect is located at approximately 4 nm depth. Through this accurate identification of defects in the gate oxide, Zhu et al. suggest the potential ferroelectric switching has for detecting small variations in signals, material properties, etc. With reliable inference accuracies, FeFETs may play an important role in the development of Analog Deep Learning as they are both simple to create and applicable to a variety of scenarios.
\\

\subsubsection{Phase Change Systems}
The method of changing phases is not specific to any one mechanism, but rather a broader method with many specific implementations, ranging from ferroelectric switching to superconductivity. Phase change devices simply employ methods of changing steady-states of a system, which can range from the temperature to the ferroelectric polarization of a material. One study suggested that phase-change memristors that could change conductance by being heated or cooled could easily be integrated with other device components\cite{pc3}. In ferroelectric-based phase change devices, a device can transition between states through the addition or removal of protons via a bias voltage\cite{pc7}, with a reported accuracy of about 89\% in MNIST digit recogntion\cite{pc1}. However, ferroelectric polarization does not always work the best in small-scale or high-temperature situations, due to issues like domain wall pinning in thin-film synaptic transistors\cite{pc4}. In such situations, other phase change systems may fair better, such as the transition of a material between crystalline and amorphous phases through the application of heat. The heat applied to the material can be controlled externally, which will impact the material's position on the crystalline-amorphous spectrum and thus change its conductance for use in a neural network\cite{bigReview}. This type of phase change system will likely prove to be the most accurate option in situations that require AI inferences in hot temperatures, such as the efficient operation of power plants, nuclear reactors, or solar farms.
\\

\noindent
\textbf{Temperature Changes:}
Ge2Sb2Te5 and Ag-and-In-doped Sb2Te memristors have come to the forefront of phase change memristors due to their gradual changes in conductance triggered by temperature changes. Similar to ionic migration memristors, these memristors have a low-resistance state and a high-resistance state. Their operation relies on controlled phase changes between a crystalline state (low-resistance) and an amorphous state (high resistance), modulated by the applied heating time. In the crystalline state, ordered vacancies with charge carriers enhance conductivity, while the amorphous phase, characterized by structural randomness, results in reduced electron mobility and higher resistance.\cite{pc3}.

Application of a positive electric field induces Joule heating in the narrow metal heater situated between the top and bottom electrodes. This process elevates the temperature of the phase change material layer beyond its melting point, causing a swift transition to the amorphous phase upon rapid quenching. Returning to the crystalline state involves maintaining the temperature above the crystallization temperature and allowing for gradual cooling.\cite{pc3}. These phase-change memristors allow for an easy implementation of neuromorphic computing in areas where temperature changes occur frequently, as they may be able to benefit the performance of a device rather than hinder it. However, this serves as a benefit to Digital Deep Learning in comparison to Analog Deep Learning, as the latter is reliant on maintaining the temperature within certain ranges to allow for the proper operation of neural networks, while the former can be operated in a much larger range of temperatures as the components are not reliant on temperature as a main factor in how they operate.
\\

\subsubsection{Metal-to-Insulator Transition}
Metal-to-insulator transition (MIT) devices are yet another application of phase change systems, as a specific material can be altered by some means (bias voltages, mechanical strain, chemical reactions, etc.) to change from a conductor to an insulator and vice versa. In between these two extreme states there exists a superposition of the two states where the material is both a conductor and an insulator. This partial conductivity can allow memristors to have different conductance values corresponding to different synaptic weights when applied to ANNs. However, MIT devices may be more prone to error than devices using other methodologies, as two studies found low value retentions\cite{mit1, mit2}. While MIT devices may not have exemplary accuracy, they may have other benefits that could outweigh the consequences, such as ease-of-use in high-temperature small-area situations.
\\

\noindent
\textbf{Mott Transion Dynamics:}
Metal-to-insulator transitions in devices as small as memristors in crossbar arrays must be done in as compact a manner as possible, allowing for the use of Mott transition dynamics. Named after the physicist Nevill Francis Mott, this transition is powered by the Coulombic effects of moving charge carriers near the device. Since the charge carriers can either attract or repel electrons, they have the ability to cancel out electric fields causing the movement of electrons in the material, which would keep the electrons in place or cause them to move, making the material an insulator or a conductor accordingly. Kumar et al.\cite{mit1} experimentally demonstrate transistor-less networks comprising third-order elements than can perform Boolean operations and find analog solutions to a computationally challenging graph-partitioning problem using neuromorphic computing powered by Mott transition dynamics. This use of metal-to-insulator transition allows for a simple implementation of neuromorphic computing but shows its drawbacks in accuracy as the study found a value retention of a little more than 1 hour. Similar to phase-change systems, metal-to-insulator transition-based devices rely on the maintenance of a steady external temperature so that the components of the crossbar array do not change conductance values simply due to environmental changes. However, in areas where the temperature is consistent, including consistently high and consistently low-temperature areas, this technology could thrive as the device's operation can easily be adjusted according to the external temperatures.
\\

\subsubsection{Superconductivity}
Superconductors may not be the first thought towards better multi-state devices, but superconductors not only provide faster electron transfers but also allow for stronger magnetic fields when they are looped. These superconducting loops can hold memory for ANNs in terms of fluxons\cite{sc2}, utilizing the quantum capabilities of certain materials such as Mott-insulating oxides\cite{sc1}. This opens up possibilities to control the fluxons using external influences, such as additional magnetic fields, which could help mimic the interconnectivity of the human brain as the device's other semiconductor loops could provide such influences. Superconductors have proven to be more useful for running ANNs themselves rather than changing the conductance values of memristors, as superconducting switches are required to transition between the analog processes of ANNs to CMOS logic in order to integrate the two architectures\cite{sc6}. While superconductors pose a radically different approach to conductance modulation, they are ultimately hard to integrate into smaller, cheaper AI products for on-the-go inferences.
\\

\noindent
\textbf{(a) Holding Memory:}
Goteti et al.\cite{sc2} present a multiterminal synapse network composed of a disordered arrangement of superconducting loops featuring Josephson junctions, in which magnetic fields are quantized into discrete fluxons, comprising microscopic circulating supercurrents, and are employed to define memory. The loops serve as traps for fluxons, and the Josephson junctions facilitate their movement between loops, creating a dynamic system with brain-like spiking information flow. Through experimental demonstrations using YBa2Cu3O7-$\delta$-based superconducting loops and Josephson junctions, the study showcases a three-loop network, exhibiting stable memory configurations of trapped flux in loops and thus influencing the rate of fluxon flow through synaptic connections. Since other parts like as transistors in logic circuits have little interference in the magnetic flux in these devices, superconductor loops coupled with Josephson Junctions have little data loss, making them a great way to implement Recurrent Neural Networks (RNNs) without compromising on inference accuracy.
\\

\noindent
\textbf{(b) Advanced Superconducting Tunnel Junctions:}
While conventional superconductors are already excellent conductors, high-transition-temperature superconductors (HTTSs) are even better conductors, operating at temperatures as low as 77 K. However, with flux-based neuromorphic computing, atomic-level control and precision is required of devices that use HTTSs in order for there to be reasonably accurate inferences in neuromorphic-computing applications. Cybart et al.\cite{sc4} propose a way to maintain such atomic-level precision in HTTSs by writing tunnel barriers directly into YBa2Cu3O(7-$\delta$) thin films using a focused helium-ion beam with a diameter of 500 pm. The researchers were able to control the barrier between the HTTPSs from states of very high conductance to states of no conductance (insulation), demonstrating the high amount of control available over the device. This can apply directly to neuromorphic computing as yet another method of conductance modulation, as the barrier's conductance can be altered during a network's backpropagation. For this implementation, however, the environmental temperature must be kept within a certain range as the HTTPSs require certain environmental conditions in order to function correctly. While this does suggest opportunities for high-speed and high-accuracy inferences, it does not help with the technology's scalability and availability at the consumer-level.
\\

\subsubsection{Spintronics}
Devices that employ spintronics involve the modulation of the physical properties of individual atoms in a material, specifically the spin of electrons. Most spintronic devices use a sandwich approach, in which one layer is fixed with a certain magnetization and the magnetization of the other layer is changed. If the electron spins of the two layers align, then the magnetization allows for less resistance and therefore greater conductance, while the opposite is true for if the two layers do not align. However, unlike digital technologies, partial conductances by the "partial" alignment of the two layers do exist, which allows for a far greater range of values each component can take on. Since electron spin is an inherent physical property in almost all materials and significantly impacts a material's conductance, it can be extremely reliable. While Spintronics may present a highly complex method that is much more difficult to test, it has tremendous potential for accurate implementations.
\\

\noindent
\textbf{(a) Hardware Acceleration:}
Due to the small scale of spintronics, parts of devices can more easily be combined to make a single integrated device, which is much more reliable and nonvolatile as issues such as the von Neumann bottleneck are removed. Nasab et al.\cite{spin5} proposed a Binarized Neural Network (BNN) harware accelerator by combining an XNOR/XOR logic circuit, a magnetic tunnel junction, and a carbon nanotube field-effect transistor into one device, integrating memory access into the operational area and thus making the device less prone to data loss and innacurate inferences. Simulating the device with BNN-based inferences, the device had an error rate of 0.0164\% for the XNOR/XOR logic circuit. While this specific implementation limits the device to BNNs, this suggests broader-scale insights for spintronics because if analog AI is accurate and precise enough to compete with devices built specifically for precise computations, it is much more likely to compete with the software-based neural networks that use digital logic circuits in their operation.
\\

\noindent
\textbf{(b) Applications in Analyzing Temporal Sequences:}
Romera et al.\cite{spin3} explore brain-inspired event binding through transient mutual synchronization of neurons, proposing its implementation in neural networks with coupled spintronic nano-oscillators for analog deep learning. The exceptional mutual synchronization ability and precise frequency tuning enable successful pattern recognition, representing progress in constructing neural networks for analog deep learning.
\par
The experiment utilizes interconnected hardware spintronic nano-oscillators, achieving synchronized activities. Involving three oscillators with individually controllable frequencies, the emitted microwave signals, initially distinct, synchronize as their frequencies approach within a mutual locking range, resulting in a more potent single peak.
\par
This synchronization phenomenon is employed in a hardware neural network with three spintronic oscillators for recognizing temporal sequences. In a fictional scenario involving a mouse encountering different cheese types, each leading to unique neuron activities, temporal sequences of three recorded spikes are used to infer the presented cheese type.
\par
Inspired by Hopfield et al.\cite{hopfield2000moment}, the researchers establish a network with neurons corresponding to the input sequence spikes, trained to recognize specific cheese types. Each spintronic oscillator is tailored to identify a particular cheese, and the network is trained to synchronize when presented with the correct spike sequence. The learning process involves adjusting parameters for the ramps leading to mutual synchronization.
\par
The network achieves a 94\% recognition rate for various cheese types, showcasing its potential for pattern recognition tasks. Comparative results with a perceptron trained on the same dataset emphasize the efficiency of this hardware-based neural network. The study suggests that spintronic oscillator synchronization holds promise for innovative and efficient applications in neuromorphic computing, particularly in recognizing temporal sequences.
\\

\subsection{Speed} 
\subsubsection{Cation Migration}
For numerous materials, the mobility of cations surpasses that of electrons, suggesting the potential for higher operational speeds in devices employing cation migration. A specific study demonstrated a switching time of less than 200 ns for a device tasked with hand-written digit classification\cite{cation2}.

\subsubsection{Anion Migration}
Devices with anions capable of higher mobility than electrons hold promise for Anion Migration, potentially achieving operational speeds comparable to electron migration-based devices. A notable study employed redox thin-film transistors (ReTFTs) to construct logic circuits using analog components, revealing a processing time of approximately 42 ns.\cite{anion2}. The use of Cation or Anion Migration-based devices can combine organization and ease-of-production with fast and accurate results, which could be used very well in consumer-level products such as on-device, offline AI inferences to keep important information regarding the inference private and more secure from interceptions.
\\

\subsubsection{Electrochemical Doping}
Although EC Doping is primarily applied in semiconductors, it holds significant potential in Analog Deep Learning due to its direct influence on a device's physical properties, facilitating faster operational speeds. An illustrative study reported a switching time of approximately 5 ns\cite{ec2}. This remarkable speed can easily compete with the double or triple-digit switching times (in ns) of ion migration implementations, and can prove useful in situations involving quick AI inferences and decision-making, such as the street driving capabilities in autonomous cars.
\\

\subsubsection{Ferroelectric Gating}
While Ferroelectric Gating may be perceived as a comparatively slower technique, it remains capable of delivering reasonable operational speeds when applied to Analog Deep Learning devices. In a specific study, a phototransistor with ferroelectric polarization exhibited a remarkable response time of under 20 $\mu$s\cite{ferroelectric2}. Additionally, another study achieved successful testing of 75 ns electrical pulses using a 5-bit FeFET synapse\cite{ferroelectric5}.
\\

\subsubsection{Phase Change Systems}
Phase Change-based devices primarily rely on Ferroelectric Tunnel Junctions (FTJs), effectively combining enhanced control over ferroelectric gating with the precision of multi-state devices. Notably, a study demonstrated that a memristor utilizing an Ag/BaTiO3/Nb:SrTiO3 ferroelectric tunnel junction (FTJ) achieved an impressive operational speed of 600 ps and accommodated 32 states (5 bits) per cell\cite{pc6}. As with any device based on electron migration, the operational speed can be further enhanced by optimizing the electrode function and increasing the carrier concentration.
\\

\subsubsection{Metal-to-Insulator Transition}
While Metal-Insulator Transition (MIT) may be perceived as slow due to the typical speed of material-level changes, its practicality and potential benefits become evident when applied to significantly smaller devices. For instance, a study demonstrated a relatively quick switching time of approximately 0.1 $\mu$s for a device designed for boolean operations\cite{mit1}. MIT is thus a strong contender for use in situations requiring simplicity but also very quick inferences.
\\

\subsubsection{Superconductivity}
Superconductors, renowned for their capacity to carry currents through charge distributions, possess significant potential for robust magnetic field generation and maintenance. The inherent speed of superconductors is advantageous in altering the values stored in magnetic fields, as rapid changes in current within the superconducting loops swiftly modulate these values. In a specific study, a device equipped with a Josephson Junction, designed for light detection, demonstrated a turn-on time of 300 ps and a turn-off time of 15 ns\cite{sc6}.
\\

\subsubsection{Spintronics}
Given that electron spin is a fundamental characteristic of materials, external control over it can be exceptionally challenging. Nevertheless, when practically achievable, this method holds the potential to yield notable benefits in operational speed. A specific study reported a switching time of 13 $\mu$s for a device designed for pattern recognition tasks\cite{spin3}.
\\

\subsection{Energy efficiency and Power Consumption}

\subsubsection{Cation/Anion Migration}
By leveraging the movement of a material's cations instead of electrons, the ion migration methods exhibit significant potential for energy efficiency in the context of Deep Learning. In a notable instance, the utilization of Electrochemical Neuromorphic Organic Devices (ENODes) enabled a study to replicate biological synapses with energy consumption of less than 10 pJ for devices spanning an area of 103 $\mu$m$^{2}$, thereby achieving reduced energy usage through a distinct mechanism as compared to memristor-based devices.\cite{cation5}. The ion migration methods offer opportunities for modular design in assembling networks from individual synaptic transistors, which are assembled from components such as electrical gating and containers with electrolytic materials. Such organized design allows for easier investigation into potential improvements for the methods over time, as opposed to more involved methods such as Superconductors and Spintronics.
\\

\subsubsection{Electrochemical Doping}
In comparison to the ionic migration methods, EC Doping involves more high-level resources as the bias voltage is not the only part of the system being changed to change conductance values. Since the material is electrochemically, the synaptic transistors that result from this method will be much more stable and reliable as the behaviors of the system will be more predictable\cite{bigReview}. For this reason, ANNs with EC Doping will likely output more accurate results than those with ion migration in a similar amount of energy usage, but the energy required on the production end will be much greater due to the need for high amounts of energy in Electrochemical Doping processes\cite{bigReview}.
\\

\subsubsection{Ferroelectric Gating}
While it might seem intuitive to assume that ferroelectric gating does not require improvements in energy efficiency due to its reliance on bias voltages, Ferroelectric Gating is, in fact, equally competent as other methods, with each having its unique specialty that appeals to specific use cases. Among all the methods discussed in this paper, Ferroelectric Gating stands out for its remarkable energy efficiency, as demonstrated by a study reporting a power consumption of 32 pJ for a 100 $\mu$m$^{2}$ device implementing reinforcement learning\cite{ferroelectric4}.
\\

\subsubsection{Phase Change Systems and Metal-to-Insulator Transition}
While the Metal-to-Insulator Transition may not appear intuitively as the most energy-efficient method, it has demonstrated remarkable efficiency when applied to Analog Deep Learning. In one study, third-order nanocircuit elements were employed in Deep Learning, resulting in a reported power consumption of approximately 2 pJ\cite{mit1}. Additionally, another study utilized a V/VOx/HfWOx/Pt Memristor within a Spiking Convolutional Neural Network, achieving an impressive energy efficiency of around 0.1 pJ. In general, Phase Change systems have potential to be very efficient methods of Analog Deep Learning, both in terms of space and energy. This efficiency comes from the fact that there is not much surface area required on a device to implement this method and the material does not need to be converted completely into a metal or an insulator to contribute to a functional neural network. This, coupled with the high speed of MIT-based inferences, make Phase Change-based networks a viable alternative to ion migration methods in situations requiring simplicity and ease-of-use.
\\

\subsubsection{Superconductivity}
While magnetization is inherently less controllable than some alternative methods, it still holds promise for providing accurate and efficient solutions to Analog Deep Learning challenges. A noteworthy study discovered that the coupling of Josephson Junctions and superconductors with Artificial Neural Networks can achieve an impressive spiking energy of less than 1 aJ\cite{sc3}, notably surpassing the energy efficiency of the human brain, which operates at approximately 10 fJ. While superconductors have remarkable energy usages when applied to resistive crossbar arrays in testing environments, they can easily be affected by external factors in consumer-level applications due to their analog nature, and may not be the most efficient in terms of energy usage for accurate inferences.
\\

\subsubsection{Spintronics}
Spintronics provides yet another avenue for efficient AI inferences through Analog Deep Learning, as magnetic properties of materials can be utilized in very small areas to make drastic differences. For example, using the magnetic spin properties of skyrmions can yield great energy and space efficiency since particles in a skyrmion gas can send random thermal signals to other particles, thereby mimicking neurons in ANNs\cite{spin1}. Additionally, impedances such as noise from the environment can be minimized by different uses of Spintronics within devices, as one study found Spintronics-based logic operations to be much more prone to noise when using spin accumulation instead of spin flow, to account for the variable electron velocities and orientations\cite{spin6}. Devices that employ this method of Analog Deep Learning can expect to be very packageable and easy to use in applications involving environmental changes, in contrast to Phase Change systems, which thrive in specific situations.

\begin{table*}[!ht]
    \centering
    \caption{Comparative performance of different representative analog deep learning methods in terms of accuracy, speed, and power consumption. Empty cell implies corresponding information is not available from the respective paper.}
    \scalebox{0.85}
    {
    \begin{tabular}{|l|l|l|l|l|l|l|}

    \hline
\textbf{Ref.} & \textbf{Method} & \textbf{Algorithms} & \textbf{Tasks} & \textbf{Accuracy} & \textbf{Switching time} & \textbf{Power consumption} \\ \hline
\cite{cation1} & Cation & RNN, RC & Sentence classification & 91\% & ~ & ~ \\ \hline
\cite{cation2} & Cation & ~ & Hand-Written Digit Classification & 90\% & $<$ 200 ns & ~ \\ \hline
\cite{cation3} & Cation & ANN, RC & Classifying Neural Firing Patterns & 86.75\%/91.8\% (training/test) & 5-20 ms & ~ \\ \hline
\cite{cation4} & Cation & ANN & MNIST digit classification & 94\% & 100 ms & $<$ 5.0 x 10$^{-10}$ J \\ \hline
\cite{cation5} & Cation & ~ & ~ & ~ & ~ & $<$ 10 pJ for 103 $\mu$m$^{2}$ devices \\ \hline
\cite{cation6} & Cation & ANN & ~ & ~ & ~ & ~ \\ \hline
\cite{anion2} & Anion & ~ & ~ & ~ & 42 ns & ~ \\ \hline
\cite{anion3} & Anion & BM & ~ & ~ & ~ & ~ \\ \hline
\cite{anion5} & Anion & CNN & MNIST Image Recognition & 96\% & ~ & ~ \\ \hline
\cite{ec1} & EC Doping & ~ & ~ & ~ & ~ & ~ \\ \hline
\cite{ferroelectric1} & Ferroelectric & ~ & ~ & ~ & 250 ns & ~ \\ \hline
\cite{ferroelectric2} & Ferroelectric & ~ & ~ & ~ & $<$ 20 $\mu$s & ~ \\ \hline
\cite{ferroelectric4} & Ferroelectric & RL & ~ & Nonlinearity (0.56/-1.23) & ~ & 32 pJ / 100 $\mu$m$^{2}$ \\ \hline
\cite{ferroelectric5} & Ferroelectric & DNN & Transient Presiach model & ~ & 75 ns & ~ \\ \hline
\cite{ferroelectric6} & Ferroelectric & ANN & Pattern Recognition & 81\% & ~ & ~ \\ \hline
\cite{pc1} & Phase Change & CNN & MNIST digit recognition & 89\% & 0.2 s & ~ \\ \hline
\cite{pc5} & Phase Change & SNN & Pattern Recognition & 100/80\% (low/high noise) & 100 ns & ~ \\ \hline
\cite{pc6} & Phase Change & ANN & MNIST digit recognition & $>$ 90\% & 600 ps & ~ \\ \hline
\cite{mit1} & MIT & ~ & ~ & ~ & 0.1 $\mu$s & 2 pJ \\ \hline
\cite{mit2} & MIT & SCNN & ~ & ~ & 1 $\mu$s & 0.1 pJ \\ \hline
\cite{sc2} & Superconductivity & ANN & ~ & ~ & ~ & ~ \\ \hline
\cite{sc3} & Superconductivity & DNN & ~ & ~ & ~ & $<$ 1 aJ \\ \hline
\cite{sc6} & Superconductivity & ~ & Detecting light & ~ & 15 ns/300 ps (off/on) & 0.18 fJ / $\mu$m$^{2}$ \\ \hline
\cite{sc7} & Superconductivity & ~ & ~ & 71\% & ~ & ~ \\ \hline
\cite{spin1} & Spintronics & ~ & Stochastic Computing Circuit & ~ & ~ & ~ \\ \hline
\cite{spin2} & Spintronics & ANN & MNIST Digit Classification & ~ & ~ & ~ \\ \hline
\cite{spin3} & Spintronics & ~ & Pattern Recognition & 94\% & 13$\mu$s & ~ \\ \hline
\cite{spin4} & Spintronics & ~ & Pattern Recognition & ~ & ~ & ~ \\ \hline
\cite{spin5} & Spintronics & CNN & ~ & logic error rate of 0.0164\% & ~ & ~ \\ \hline
\end{tabular}
    }
\end{table*}

\section{Discussion}
While Analog Deep Learning is an emerging field that has proven to have tremendous potential in further developing AI, it is not completely error-free. Due to the nature of analog processing, the accuracy of AI inferences can be compromised. Analog systems are very prone to noise, which can be a limiting factor in a device's performance. Due to accuracy and noise considerations, Analog Deep Learning may not be suitable for many high-stakes applications, such as identifying threats by a camera and computer vision. However, for consumer-level applications such as recognizing and pulling out text from images, the use of analog deep learning is likely to be very beneficial.

\subsection{Cost} 
While some methods of Analog Deep Learning, such as Spintronics and Superconductivity, may require expensive materials or resources, some methods have proven to be very cost efficient for use in AI applications. For example, Ferroelectric Gating only requires the use of bias voltages across a material to induce its conductance, which is likely to require relatively inexpensive resources such as a voltage source, the material itself, and the supporting electronics.

\subsection{Scalability} 
The Analog Deep Learning methods that are likely to be very cost efficient are also likely to be used in mass integration and production. While successful and beneficial emulation of a single neural network can be very efficiently done by an analog device, emulation of multiple neural networks in one cohesive device is likely to make AI inferences much stronger while keeping expenses at bay. Integration of multiple neural networks may prove to be a better solution than creating one very large neural network in applications such as text writing programs, where parallelism can significantly boost efficiency.
\par
Depending on the applications, some Deep Learning algorithms or hardware implementation methods may be more suitable. For lower-cost, consumer-level applications such as personal assistants, Cation Migration is likely to be the best implementation method because of its simplicity. Personal assistants may also benefit from the use of RNNs, as RNNs are likely to incorporate aspects of memory better than other algorithms such as ANNs. Other applications such as self-driving systems are likely to benefit with the use of CNNs due to the high amount of spatial awareness embedded in CNN-based devices. Self-driving systems are likely to benefit from ion migration or material-level change implementation types due to their simplicity and low cost of materials, as methods such as superconductivity and spintronics may unnecessarily drive up device prices.
\par 
The simplicity of a crossbar array with ion migration or charge distribution-based devices cannot be understated. With such an approach we can shrink down neural networks and directly model them with hardware, albeit with a small tradeoff of accuracy depending on the situation. In the present day, AI is very much clustered in areas with access to key resources and computing power, but with Analog Deep Learning we may be able to place immense power in the hands of day-to-day consumers.

\section{Conclusion}
In this review, we introduced the increasingly popular field of Analog Deep Learning and discussed its applications, current progress, and future potential. We described a taxonomy for classifying implementations of Analog Deep Learning, and analyzed each of the eight primary methods. We then proceeded to comparing the methods and evaluating their scalability and future implications.
\par 
From this review, we conclude that Analog Deep Learning should be further studied, especially for lower-cost or consumer-level applications such as personal assistants or driving systems. With methods such as cation migration and spintronics which are relatively easy to implement and/or have adequate accuracy and efficiency, Analog Deep Learning is likely to allow Artificial Intelligence to reach new heights.

\bibliographystyle{IEEEtran}
\bibliography{ref}

\EOD
\end{document}